\pdfoutput=1

\documentclass[11pt]{article}

\usepackage{naacl2021}

\usepackage{times}
\usepackage{latexsym}

\usepackage[T1]{fontenc}

\usepackage[utf8]{inputenc}

\usepackage{microtype}%

\usepackage{comment}

\usepackage{color}
\usepackage{graphicx}
\usepackage{tikz}
\usepackage{amsmath}

\usepackage{enumitem}
\usepackage{booktabs}
\usepackage{array,multirow}
\usepackage{latexsym}
\usepackage{multicol}
\usepackage{colortbl}
\usepackage{makecell}
\usepackage{arydshln}
\usepackage{tabu}
\usepackage{wrapfig}

\usepackage[greek,english]{babel}

\title{Scalar Adjective Identification and  Multilingual Ranking}

\author{Aina Gar\'i Soler \\
  Université Paris-Saclay \\ 
  CNRS, LISN \\ 
  91400, Orsay, France \\
  \texttt{aina.gari@limsi.fr} \\\And
  Marianna Apidianaki \\
  Department of Digital Humanities \\
  University of Helsinki \\
  Helsinki, Finland \\
  \texttt{marianna.apidianaki@helsinki.fi} \\}

\begin{document}
\maketitle

\begin{abstract}
The intensity relationship that holds between scalar adjectives (e.g., {\it nice} < {\it great} < {\it wonderful}) is highly relevant for natural language inference and common-sense reasoning. 
Previous research on scalar adjective ranking has 
focused on English, mainly due to the availability of 
datasets for evaluation. We introduce a new multilingual dataset in order to promote research on scalar adjectives 
in new languages. We perform a series of experiments 
and set performance baselines on this dataset, using 
monolingual and multilingual contextual language  
models. 
Additionally, we introduce a new binary classification task for English scalar adjective identification  
which examines the models'  
ability to distinguish scalar from relational adjectives. 
We probe contextualised representations and report baseline results for future comparison on this task.
\end{abstract}

\section{Introduction}

Scalar adjectives relate the entities they modify  
to specific  
positions on the evoked scale 
(e.g., {\sc goodness}, {\sc temperature}, {\sc size}):  A \textit{wonderful view} is nicer than  
a \textit{good view}, and one would probably prefer a {\it delicious} to a {\it tasty  meal}. But not all adjectives express intensity or degree. 
Relational adjectives 
are derived from nouns (e.g., {\it wood} $\rightarrow$ {\it wooden}, {\it chemistry} $\rightarrow$ {\it chemical}), have no antonyms 
and serve to classify 
nouns (e.g., {\it a wooden table}, {\it a chemical substance}) 
\cite{mcnally2004relational}.  
The distinction between scalar and relational adjectives is an important one. Identifying adjectives that express intensity can serve to assess the emotional tone 
of a given text, as opposed to words that mostly contribute to its descriptive content. 
Additionally, estimating the intensity of a scalar adjective 
is useful  
for textual entailment 
(\textit{wonderful}  $\models$  \textit{good} but {\it good}  $\not\models$ \textit{wonderful}), 
product review analysis and recommendation systems,
emotional chatbots and question answering \cite{demarneffe:10}.

\begin{table}[t!]
    \centering
    \scalebox{0.8}{
    \begin{tabular}{c|p{8.2cm}}
     &  \parbox{8.2cm}{\begin{center}{\sc DeMelo}\end{center}} \\
          \cdashline{2-2}
         {\sc en} & dim <	gloomy < dark < black \\
         {\sc fr} & terne < sombre < foncé <	noir \\
         {\sc es} & sombrío < tenebroso < oscuro < negro \\
         {\sc el} & \textgreek{αμυδρός} $\vert\vert$ \textgreek{αχνός} < \textgreek{μουντός}	< \textgreek{σκοτεινός}< \textgreek{μαύρος} \\ 
         \cdashline{2-2}
          & \parbox{8.2cm}{\vspace{3mm}\begin{center}{\sc Wilkinson}\end{center}} \\
         \cdashline{2-2}
         {\sc en} & bad < awful <	terrible < horrible \\
         {\sc fr} & mauvais < affreux < terrible <	horrible \\
         {\sc es} & malo < terrible < horrible < horroroso\\
         {\sc el} & \textgreek{κακός}	< \textgreek{απαίσιος }< \textgreek{τρομερός} < \textgreek{φρικτός} \\
        \cdashline{2-2}
        
    \end{tabular}}
    \caption{Example translations from each dataset. ``$\vert\vert$'' indicates 
    adjectives at 
    the same intensity level (ties).}
    \label{tab:example_translations}
\end{table}

Work on scalar adjectives has 
until now evolved around pre-compiled datasets \cite{demelo:13,taboada-etal-2011-lexicon,wilkinson2016gold,cocos-2018-learning}. Reliance on external resources 
 has also restricted research to English, and has led to the prevalence of pattern-based and lexicon-based approaches. 
Recently, \newcite{gari2020puntacana} showed that BERT  representations \cite{devlin2019bert} encode intensity relationships between English scalar adjectives, paving the way for applying contextualised representations to intensity detection in other languages.\footnote{\citet{demelo:13} discuss the possibility of a  pattern-based multilingual approach which would require the translation of English patterns (e.g., ``X but not Y'') into other languages.
} 

In our work, we explicitly address the scalar adjective identification task, overlooked until now due to the focus on 
pre-compiled resources. 
We furthermore propose to extend scalar adjective ranking to new  languages. We make available 
two new benchmark datasets for scalar adjective identification and multilingual ranking: (a) {\sc scal-rel}, a balanced dataset of relational 
and scalar adjectives 
which can serve to probe model representations for scalar adjective identification; 
and (b) {\sc multi-scale}, a scalar adjective dataset in  
French, Spanish and Greek. In order to test contextual models 
on these two tasks,  
the adjectives need to be seen in sentential context. We thus provide, alongside the datasets, sets of sentences that can be used to extract contextualised 
representations in order to promote model comparability. 
We conduct experiments and report results obtained with simple baselines and state-of-the-art monolingual and multilingual models on these new benchmarks, opening 
 up avenues for research 
 on sentiment analysis and emotion detection in different languages.\footnote{Our code and data are available at \url{https://github.com/ainagari/scalar_adjs}.}

\section{The Datasets}

\subsection{The MULTI-SCALE Dataset} \label{sec:multi_scale}

We translate 
two English scalar adjective 
datasets into French, Spanish and Greek: 
{\sc deMelo}  
consists of 87 hand crafted half-scales\footnote{
A full scale 
(e.g., \{\textit{hideous} > 
\textit{ugly},  
\textit{pretty} < 
\textit{beautiful} < 
\textit{gorgeous}\} can be split into two half scales 
which contain antonyms,  
often 
expressing 
different polarity \{\textit{hideous} 
> \textit{ugly}\} and 
\{\textit{pretty} 
< \textit{beautiful} 
< \textit{gorgeous}\}.} \cite{demelo:13}  
 and {\sc Wilkinson}  
contains 12 full scales \cite{wilkinson2016gold}. We use the partitioning of  {\sc Wilkinson} into 21 half-scales 
proposed by 
\citet{cocos-2018-learning}. 
In what follows, we use the term ``scale'' to refer to half-scales. 

The two translators have (near-)native 
proficiency in 
each language. 
They were shown  the adjectives in the context of a scale.  
This context narrows down the possible translations for polysemous adjectives to the ones that express 
the  meaning 
described inside the scale. For example, the 
Spanish translations proposed for the adjective {\it hot} in the scales  \{\textit{warm} < {\it {\underline {hot}}}\} and \{\textit{flavorful} < \textit{zesty} < \textit{{\underline {hot}}} || \textit{spicy}\} are \textit{caliente} and \textit{picante}, respectively. 
Additionally, the translators 
were instructed to preserve the number of words in the original scales when possible.
In some cases, however, 
they proposed alternative translations 
for English words, or none if 
an adequate translation could not be found. As a result, the translated datasets have a different number of words and ties.
Table \ref{tab:example_translations} shows examples of  original English scales and their  French, Spanish and Greek translations. 
Table \ref{tab:dataset_stats} contains  
statistics on the 
composition of the translated 
datasets.

\begin{table}[]
    \centering
    \scalebox{0.95}{
    \begin{tabular}{cc|cc}
          &  & \makecell{\# unordered  
          pairs} & 
          \makecell{ \# adjectives} \\
          \hline
          \parbox[t]{2mm}{\multirow{4}{*}{\rotatebox[origin=c]{90}{\sc demelo}}}
          
          & {\sc en} & 548 (524) & 339 (293) \\
          & {\sc fr} & 590 (567) & 350 (303)  \\
          & {\sc es} & 448 (431) & 313 (275) \\
           & {\sc el} & 557 (535) & 342 (295) \\
           
           \hline
           \parbox[t]{2mm}{\multirow{4}{*}{\rotatebox[origin=c]{90}{\sc wilkinson}}} & {\sc en} & 61 (61) & 59 (58) \\
          & {\sc fr} & 67 (67) & 61 (60) \\
          & {\sc es} & 59 (59)  & 58 (56) \\
          & {\sc el} & 68 (68) & 61 (58) \\
          
    \end{tabular}
    }
    \caption{Composition 
    of the translated datasets. In parentheses, we  give the number of unique adjectives and pairs.
    } 
    \label{tab:dataset_stats}
\end{table}


In order to test contextual models on the ranking task, we collect 
sentences containing the adjectives 
from 
OSCAR \cite{suarez2019asynchronous}, 
a multilingual corpus 
derived from CommonCrawl. 
French, Spanish and Greek are morphologically rich languages where adjectives need to agree 
with the noun they modify.
In order to keep the 
method 
resource-light, 
 we gather sentences that contain the adjectives in their unmarked form. 

For each scale $s$, 
we randomly select ten sentences from OSCAR where  
adjectives from $s$ occur. 
Then, 
we generate additional 
sentences through lexical substitution. Specifically, 
for every sentence (context) $c$ that contains an adjective 
$a_i$ from  scale $s$, we replace $a_i$ with 
$\forall$ $a_j$ $\in s$ where  $j = {1...|s|}$ and $j \neq i$. 
This process results in a total of $|s|$ * 10 sentences per scale and ensures that $\forall$ $a$ $\in s$  is seen in the same ten contexts. 
For English, 
we use the ukWaC-Random set of sentences compiled  
by \citet{gari2020puntacana} which contains sentences randomly collected 
from the ukWaC corpus 
\cite{baroni2009wacky}. 

\subsection{
The SCAL-REL Dataset} \label{sec:scal_rel}


{\sc scal-rel} contains 
scalar adjectives 
from the {\sc deMelo}, {\sc Wilkinson} and {\sc Crowd} \cite{cocos-2018-learning} datasets 
(i.e. 79 additional half-scales compared to {\sc multi-scale}). 
We use all unique scalar adjectives in the 
datasets 
 (443 
 in total), and subsample 
 the same number of relational adjectives,   
 which are labelled with the pertainym relationship in WordNet \cite{Fellbaum1998}. There are 4,316 unique such 
 adjectives in WordNet,   
including many rare or highly technical terms  
(e.g., \textit{birefringent}, \textit{anaphylactic}).\footnote{Note that the WordNet annotation 
does not cover all pertainyms in English (for example, frequent words such as \textit{ironic} or \textit{seasonal} are not marked with this relation).} 
Scalar adjectives in our datasets are 
much more frequent 
than these relational adjectives; 
their average frequency 
in Google Ngrams \citep{brants2006web} is 27M and 1.6M, respectively. 
We balance 
the relational adjectives set by frequency, 
by subsampling 
222 frequent and 221 rare 
adjectives. We use 
the mean 
frequency of the 4,316  
relational adjectives 
in  Google Ngrams as a  threshold.
\footnote{Nine scalar adjectives from our datasets are also annotated as pertainyms in WordNet (e.g.,  \textit{skinny, microscopic}) because they are denominal. We consider these adjectives to be scalar for our purposes since they clearly belong to intensity scales.}
We propose a train/dev/test split of the 
{\sc scal-rel} dataset (65/10/25\%),  
observing a balance between the two classes (scalar and relational) in each set. 
To obtain 
contextualised representations, 
we collect 
for each relational adjective 
ten random sentences from ukWaC. For scalar adjectives, we use the ukWaC-Random set of sentences (cf. Section \ref{sec:multi_scale}). 

\section{Multilingual Scalar Adjective Ranking}
\label{vectorsub}

\subsection{Methodology} \label{sec:methodology}

\noindent {\bf Models} We conduct experiments with state-of-the-art contextual language models 
and several baselines on the {\sc multi-scale} dataset. 
We 
use the pre-trained {\tt cased} and {\tt uncased}  multilingual BERT model \cite{devlin2019bert} and  
report results of the best variant 
for each language. 
We also report results 
obtained with four 
monolingual models: {\tt bert-\allowbreak base-\allowbreak uncased} \cite{devlin2019bert}, 
{\tt flaubert\_\allowbreak base\_\allowbreak uncased} \cite{le2020flaubert}, 
{\tt bert-\allowbreak base-\allowbreak spanish-\allowbreak wwm-\allowbreak uncased} 
\citep{CaneteCFP2020}, and {\tt bert-\allowbreak base-\allowbreak greek-\allowbreak uncased-\allowbreak v1} \cite{koutsikakis2020greek}. 
We compare to results 
obtained using 
fastText static  embeddings 
in each language 
\cite{grave2018learning}.

For a scale $s$, we feed the corresponding set of sentences 
to a model and extract the contextualised representations for 
$\forall$ $a \in s$ 
from every layer. 
When an adjective is split into multiple BPE units,   
we average 
the representations of all wordpieces (we call this approach ``WP'')   
or 
all pieces but the last one (``WP-1'').  
The intuition behind excluding the last WP 
is that the ending 
of a word often corresponds to a suffix 
with 
morphological information. 
\vspace{2mm}

\noindent {\bf The {\sc diffvec} method}  We apply the  
 adjective ranking
 method proposed by \citet{gari2020puntacana}  to our dataset, which relies on 
 an intensity vector (called $\overrightarrow{dVec}$) built from   
 BERT representations. 
 The method yields state-of-the art results with very little data;  this makes it easily adaptable 
 to new 
 languages. We build a sentence specific  intensity representation ($\overrightarrow{dVec}$) by subtracting the vector of a 
mild  intensity adjective,  $a_{mild}$ 
(e.g.,  \textit{smart}), from that of  $a_{ext}$, 
an extreme 
adjective 
on the same scale (e.g., {\it brilliant}) in the same context. 
We create a 
$dVec$ representation from every sentence available for 
these two reference 
adjectives,  
and average them to obtain the global  $\overrightarrow{dVec}$ for that pair. 
\citet{gari2020puntacana} showed that  
a single positive adjective pair ({\sc diffvec-1} $(+)$) 
is
enough for obtaining highly competitive  
results in English.  
We apply this method to  
the other languages using  
the translations of a  positive English ($a_{mild}$, 
$a_{ext}$) pair 
from the {\sc Crowd} dataset: \textit{perfect-good}.\footnote{{\sc fr}: parfait-bon, {\sc es:} perfecto-bueno, {\sc el}: \textgreek{τέλειος}-\textgreek{καλός}.}  

Additionally, we learn  
two dataset specific representations: one by averaging the $\overrightarrow{dVec}$'s of all 
($a_{ext}$, $a_{mild}$) 
pairs in {\sc Wilkinson} that do not appear in {\sc deMelo} ({\sc diffvec-wk}), and 
another one from 
pairs in {\sc deMelo} that are not in {\sc Wilkinson}  
({\sc diffvec-dm}).  
We rank adjectives in a scale by their cosine similarity to each $\overrightarrow{dVec}$:  
The higher the similarity, the  more intense the adjective is. 

\vspace{2mm}

\noindent \textbf{Baselines} We 
compare our results to a frequency and a polysemy baseline ({\sc freq} and {\sc sense}). 
These baselines rely on the assumption   that low intensity words (e.g., \textit{nice, old}) are 
more frequent and polysemous than  their extreme counterparts (\textit{e.g., awesome, ancient}). Extreme adjectives often limit 
the denotation of a noun to a smaller class of referents than 
mild intensity adjectives \citep{geurts2010quantity}. For example, an ``awesome view'' is more rare than a ``nice view''. This assumption  has  been confirmed 
for English in  \citet{gari2020puntacana}. {\sc freq} orders words in a scale according to their frequency: 
Words with higher frequency have lower intensity.  
Given the strong correlation between word frequency and number of senses \citep{Zipf1945}, 
we also expect highly polysemous words 
(which are generally more frequent) to have lower intensity. 
This is captured by the {\sc sense} baseline which orders the words according to their number of senses: Words with more senses have lower intensity.

\begin{table*}[ht]
    \centering
    \scalebox{0.81}{
    \begin{tabular}{cc | ccc | ccc | ccc | ccc}
    & \multicolumn{1}{c}{} & \multicolumn{3}{c}{{\sc en}} & \multicolumn{3}{c}{{\sc fr}} & \multicolumn{3}{c}{{\sc es}} & \multicolumn{3}{c}{{\sc el}} \\
  
         & \multicolumn{1}{c}{} & \multicolumn{3}{c}{{\bf Mono WP-1}} & \multicolumn{3}{c}{{\bf Mono WP-1}}  & \multicolumn{3}{c}{{\bf Mono WP-1}}  & \multicolumn{3}{c}{{\bf Mono WP-1}}\\
         \hline
          & & {\sc p-acc} & $\tau$ & $\rho_{avg}$ & {\sc p-acc} & $\tau$  & $\rho_{avg}$ & {\sc p-acc} & $\tau$ & $\rho_{avg}$ & {\sc p-acc} & $\tau$  & $\rho_{avg}$ \\
          \hline
        
         \parbox[t]{2mm}{\multirow{2}{*}{\rotatebox[origin=c]{90}{\sc DM 
         }}} & {\sc dv}-1 $(+)$ & \textbf{.651}$_{9}$ & \textbf{.435}$_{9}$ & \textbf{.496}$_{9}$ &  \textbf{.610}$_{3}$ & \textbf{.369}$_{3}$ & \textbf{.396}$_{3}$ & .658$_{9}$ & .381$_{9}$ & \textbf{.407}$_{9}$ & .564$_{2}$ & .238$_{1}$ & .271$_{2}$ \\

         & {\sc dv-wk} & .586$_{6}$ & .267$_{6}$ & .300$_{6}$ & .515$_{1}$ & .167$_{1}$ & .166$_{7}$ & \textbf{.670}$_{7}$ & \textbf{.404}$_{7}$ & \textbf{.407}$_{7}$ & .589$_{2}$ & .294$_{2}$ & .325$_{2}$ \\ 
         \hline
         \parbox[t]{2mm}{\multirow{2}{*}{\rotatebox[origin=c]{90}{\sc WK}}} 
         & {\sc dv}-1 $(+)$ & .852$_{1}$ & .705$_{1}$ & .802$_{1}$ & .612$_{6}$ & .257$_{6}$ & .215$_{6}$ & \textbf{.814}$_{7}$ & \textbf{.627}$_{7}$ & \textbf{.803}$_{9}$ & .618$_{8}$ & .282$_{8}$ & .256$_{8}$\\

         & {\sc dv-dm} & \textbf{.918}$_{10}$ & \textbf{.836}$_{10}$ & \textbf{.859}$_{10}$ & .642$_{7}$ & .322$_{2}$ & .392$_{2}$ & .780$_{6}$ & .559$_{6}$ & .684$_{6}$ & \textbf{.750}$_{10}$ & \textbf{.564}$_{10}$ & \textbf{.586}$_{10}$ \\ 
         \hline
         & \multicolumn{1}{c}{} &    
   \multicolumn{3}{|c}{{\bf   Multi WP-1}} & \multicolumn{3}{|c}{{\bf Multi WP}} & \multicolumn{3}{|c}{{\bf Multi WP}} & \multicolumn{3}{|c}{{\bf Multi (unc) WP}}\\
         \hline
          \parbox[t]{2mm}{\multirow{2}{*}{\rotatebox[origin=c]{90}{\sc DM}}} 
          & {\sc dv}-1 $(+)$ & .609$_{4}$ & .346$_{4}$ & .389$_{4}$ & .559$_{7}$  & .260$_{7}$ & .311$_{7}$ &  .614$_{3}$ & .291$_{3}$ & .268$_{5}$ & .517$_{9}$ & .139$_{9}$ & .163$_{9}$\\

         & {\sc dv-wk} & .544$_{3}$ & .208$_{3}$ & .241$_{4}$ & .517$_{10}$ & .170$_{10}$ & .179$_{10}$ & .618$_{12}$ & .301$_{12}$ & .303$_{12}$ & .539$_{9}$ & .181$_{9}$ & .207$_{9}$\\
         
          \hline
\parbox[t]{2mm}{\multirow{2}{*}{\rotatebox[origin=c]{90}{\sc WK 
}}} & {\sc dv}-1 $(+)$ & .836$_{6}$ & .672$_{6}$ & .717$_{6}$ & .672$_{3}$ & .382$_{3}$ & .380$_{3}$ & .797$_{3}$ & .593$_{3}$ & .639$_{3}$ & .662$_{10}$ & .388$_{9}$ & .423$_{9}$\\

         & {\sc dv-dm} & .836$_{7}$ & .672$_{7}$ & .766$_{7}$ & \textbf{.701}$_{6}$ & \textbf{.441}$_{6}$ & \textbf{.476}$_{2}$ & .695$_{10}$ & .390$_{10}$ & .511$_{10}$ & .691$_{5}$ & .447$_{5}$ & .502$_{5}$ \vspace{2mm}\\ \hline 
         
                   &  & \multicolumn{12}{c}{{\bf Static models and baselines}}  \\ \hline
          \parbox[t]{2mm}{\multirow{4}{*}{\rotatebox[origin=c]{90}{\sc DM}}} & {\sc dv}-1 $(+)$ &.637 & .407 & .458 & .573 & .288 & .275 & .656 & .383 & \textbf{.421} & .575 & .266 & .273 \\ 
          & {\sc dv-wk} & .599 & .330 & .406 & .454 & .033 & -.006 & .616 & .298 & .315 & .549 & .205 & .217\\
          \cdashline{2-14}
         & {\sc freq} & .575 & .271 & .283 &  .602 & .346 & .345 & .585 & .227 & .239 & \textbf{.596} & \textbf{.306} & \textbf{.334}\\
         & {\sc sense} & .493 & .163 & .165 & .512 & .229 & .185 & .516 & .139 & .151 & - & - & -\\ \hline          \parbox[t]{2mm}{\multirow{4}{*}{\rotatebox[origin=c]{90}{\sc WK}}} & {\sc dv}-1 $(+)$ & .787 & .574 & .663 & .582 & .197 & .152 & .695 & .390 & .603 & .706 & .464 & .566\\
     
         & {\sc dv-dm} & .852 & .705 & .783 & .642 & .325 & .280 & .712 & .424 & .547 & .691 & .447 & .451\\
          \cdashline{2-14}
         & {\sc freq} & .754 & .508 & .517 & .567 & .167 & .148 & .576 & .153 & .382 & .676 & .417 & .427\\
         & {\sc sense} & .721 & .586 & .575 & .567 & .255 & .340 & .644 & .411 & .456 
         & - & - & -\\
     
    \end{tabular}}
    \caption{Results of the {\sc diffvec} ({\sc dv}) method 
    with 
    monolingual (Mono) and multilingual (Multi) contextual models. 
    Comparison to static embeddings and baselines per language. 
    Subscripts denote the best layer. 
    The best result obtained for each  dataset in each language is indicated in boldface. 
    For all languages but Greek, the multilingual model is cased.}
    \label{tab:monomulti}
\end{table*}

Frequency is taken from Google Ngrams 
for English, and from 
OSCAR for the other three languages. 
The number of senses is retrieved from WordNet for English, and from BabelNet \cite{NavigliPonzetto:12aij} for Spanish and French.
\footnote{We omit Named Entities from BabelNet entries (e.g., names of TV shows or 
locations).}  
For 
adjectives that are not present in BabelNet,
we use a default value which corresponds to the average number of senses for adjectives in the 
dataset ({\sc deMelo} or {\sc Wilkinson}) for which this information is available. 
We  
omit the {\sc sense} baseline for Greek due to low coverage.
\footnote{Only 47\% of the Greek adjectives have a BabelNet entry, compared to 95.7\% and 88.9\% for Spanish and French. All English adjectives are present in WordNet.} 

\subsection{Evaluation}


We use 
evaluation metrics traditionally used 
for ranking evaluation  
\cite{demelo:13,cocos-2018-learning}:
Pairwise accuracy ({\sc p-acc}), Kendall's $\tau$ and Spearman's $\rho$. 
Results on this task are given in Table \ref{tab:monomulti}.
Monolingual models perform consistently better than the multilingual model,  except for French. We report the best wordpiece approach 
for each model: 
WP-1 
works better with all monolingual models and the multilingual model for English. Using all wordpieces (WP) is a better choice 
for the multilingual model 
in 
other languages. 
We believe the lower performance of 
WP-1 in these settings 
to be due to the fact that the multilingual BPE vocabulary is mostly English-driven; this naturally results in highly arbitrary partitionings in these languages (e.g., {\sc es}: \textit{fant\'astico} $\rightarrow$ fant-\'astico; {\sc el}: \textit{\textgreek{γιγάντιος}} ({\it gigantic})$\rightarrow$\textgreek{γ-ι-γ-άν-τιος}). Tokenisers of the monolingual models instead  tend to split words in a way that more closely reflects  
the morphology of the language (e.g., {\sc es}: \textit{fant\'astico} $\rightarrow$ fant\'as-tico; {\sc el}:  \textit{\textgreek{γιγάντιος}}$\rightarrow$\textgreek{γιγά-ντι-ος}.
Detailed results 
are found in Appendix \ref{app:comparison_wp_selection}.
We observe that {\sc diffvec-1} $(+)$ yields comparable and sometimes better results than {\sc diffvec-dm} and {\sc diffvec-wk},  
which are built from multiple 
pairs. This is important especially in the  multilingual setting, since it shows that just one pair of adjectives 
is enough for obtaining 
good  results in a new language.  
The best layer varies across models and configurations. The monolingual French and Greek models generally obtain  best results in earlier layers. A similar behaviour is observed for   
the multilingual model for English to some extent, whereas  
for the other models performance improves 
in the upper half of the Transformer network (layers 6-12). This shows that the semantic information relevant for adjective ranking is not situated at the same level of the Transformer in different languages. We plan to investigate this finding further in future work. 
The lower results in French can be due to 
the higher amount of ties present in 
the 
datasets compared to other languages.\footnote{58\% of the French  {\sc deMelo} scales  contain a tie, 
compared to 45\% in English.} 
The baselines obtain competitive results 
showing that the underlying linguistic intuitions 
hold across languages.
The best models beat the baselines in all configurations except for Greek on the {\sc deMelo} dataset, where {\sc freq} and static embeddings obtain higher results. Overall,  
results are lower than those reported  for English, 
which shows that there is room for improvement in new  languages.


\section{Scalar Adjective Identification}


\begin{figure}
    \centering
    \scalebox{0.75}{
    \includegraphics[width=\columnwidth]{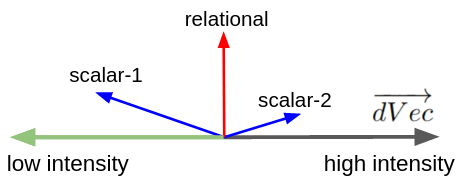}}
    \caption{Illustration of two scalar adjectives that are close to  $\overrightarrow{dVec}$ and to its opposite (which represents low intensity). The red vector describes a relational adjective that is perpendicular to $\overrightarrow{dVec}$. 
    }
    \label{fig:plot}
    
\end{figure}

For each English adjective in the {\sc scal-rel} dataset, we generate a representation 
from the available ten sentences (cf. Section \ref{sec:scal_rel}) using the {\tt bert-base-uncased} model 
(with 
WP and WP-1). 
We experiment with a 
simple logistic regression classifier that uses the 
averaged representation for an adjective ({\sc adj-rep}) 
as input and predicts 
whether it is scalar or relational. 
We also 
apply the {\sc diffvec-1} $(+)$ method to this task and measure 
how intense an adjective is by calculating its cosine with $\overrightarrow{dVec}$. 
The absolute value of the cosine 
indicates how clearly an adjective 
encodes the notion of intensity. 
In Figure \ref{fig:plot}, we show two scalar adjective vectors with negative and positive cosine similarity to $\overrightarrow{dVec}$, and another vector that is perpendicular to $\overrightarrow{dVec}$, i.e. describing a relational adjective for which the notion of intensity does not apply.\footnote{To draw a parallel with 
gender debiasing,  
this value would reveal words' bias in the gender direction \cite{Bolukbasi-NIPS2016}, regardless of the gender (male or female).} 
We train a logistic regression model to find a cosine threshold separating scalar from relational adjectives ({\sc dv}-1 $(+)$). Finally, we also use as a feature 
the cosine similarity of the adjective representation to  the vector of ``\textit{good}'', which we consider as a prototypical scalar adjective ({\sc proto-sim}). 

The best 
BERT layer 
is selected based on the 
accuracy obtained on the development set. We report accuracy on the test set. 
The baseline classifiers 
only use frequency ({\sc freq}) and polysemy ({\sc sense}) as 
features.
We use these baselines on {\sc scal-rel} because the WordNet pertainyms included in the dataset are rarer than the scalar adjectives. 
The intuition behind the {\sc sense} baseline  
explained in  Section \ref{sec:methodology} also applies here. 


\begin{table}[]
    \centering
    \scalebox{1}{
    \begin{tabular}{l|cc}
        \multirow{2}{*}{Method} & \multicolumn{2}{c}{Accuracy} \\
        & WP & WP-1 \\
        \hline
         
         {\sc adj-rep} (BERT) & \textbf{0.946}$_{9}$ & 0.942$_{9}$  \\
         {\sc proto-sim} & 0.888$_{11}$ & 0.902$_{10}$ \\
         {\sc dv-1} $(+)$ & 0.549$_{2}$ & 0.545$_{2}$ \\ \cdashline{1-3}
         {\sc adj-rep} (fastText) & \multicolumn{2}{c}{0.929}\\
         {\sc freq} & \multicolumn{2}{c}{0.669}\\
         {\sc sense} & \multicolumn{2}{c}{0.714} \\
    \end{tabular}
    }
    \caption{Classification results on the {\sc scal-rel} dataset.} 
    \label{tab:relational_results}
\end{table}

Results on this task are given in Table \ref{tab:relational_results}. 
The classifier that relies on {\sc adj-rep BERT} 
representations 
can distinguish the two types  of adjectives 
with very high accuracy 
(0.946), closely followed by fastText embeddings (0.929). 
The {\sc dv}-1 $(+)$ method
 does not perform as well as the classifier based on 
 {\sc adj-rep}, 
which is not surprising since 
it relies on a single feature (the absolute value of the cosine between $\overrightarrow{dVec}$ and {\sc adj-rep}). Comparing {\sc adj-rep}  
to a typical scalar word ({\sc proto-sim}) yields better results than {\sc dv-1 $(+)$}. 
The {\sc sense} and {\sc freq} baselines can capture the distinction to some extent. 
Relational adjectives in our training set are less frequent and have 
fewer senses on average (2.59) 
than scalar adjectives (5.30). 
A closer look at the errors of the best model 
reveals that these concern tricky cases: 
One of the four misclassified scalar adjectives is derived from a noun (\textit{microscopic}), whilst five out of eight wrongly classified relational adjectives can have a scalar interpretation (e.g., \textit{sympathetic, imperative}).
Overall, supervised models obtain very good results on this task. {\sc scal-rel} will enable research on unsupervised methods that could be used in other languages.


\section{Conclusion}

We propose a new multilingual benchmark 
for scalar adjective ranking, and set performance baselines on it using 
monolingual and multilingual 
contextual language model representations. 
Our results show that adjective intensity information is present in the contextualised representations in the studied languages. 
We also propose a new classification task and a 
dataset that can serve  as a benchmark to estimate the models' capability to identify scalar adjectives when relevant datasets are not available. We make our datasets and sentence contexts 
available to promote future research on scalar adjectives detection and analysis in different languages. 

\section*{Acknowledgements}


\setlength\intextsep{0mm}

\begin{wrapfigure}[]{l}{0pt}

\includegraphics[scale=0.3]{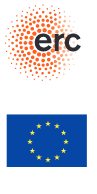}
\end{wrapfigure}

This work has been supported by the French National Research Agency under project ANR-16-CE33-0013. The work is also part of the FoTran project, funded by the European Research Council (ERC) under the European Union’s Horizon 2020 research and innovation programme (grant agreement \textnumero ~771113). We thank the anonymous reviewers for their 
valuable suggestions.

\bibliographystyle{acl_natbib}
\bibliography{anthology,eacl2021}

\begin{thebibliography}{20}
\expandafter\ifx\csname natexlab\endcsname\relax\def\natexlab#1{#1}\fi

\bibitem[{Baroni et~al.(2009)Baroni, Bernardini, Ferraresi, and
  Zanchetta}]{baroni2009wacky}
Marco Baroni, Silvia Bernardini, Adriano Ferraresi, and Eros Zanchetta. 2009.
\newblock \href
  {https://wacky.sslmit.unibo.it/lib/exe/fetch.php?media=papers:wacky_2008.pdf}
  {{The WaCky wide web: a collection of very large linguistically processed
  web-crawled corpora}}.
\newblock \emph{Journal of Language Resources and Evaluation}, 43(3):209--226.

\bibitem[{Bolukbasi et~al.(2016)Bolukbasi, Chang, Zou, Saligrama, and
  Kalai}]{Bolukbasi-NIPS2016}
Tolga Bolukbasi, Kai-Wei Chang, James~Y Zou, Venkatesh Saligrama, and Adam~T
  Kalai. 2016.
\newblock \href
  {http://papers.nips.cc/paper/6228-man-is-to-computer-programmer-as-woman-is-to-homemaker-debiasing-word-embeddings.pdf}
  {{Man is to Computer Programmer as Woman is to Homemaker? Debiasing Word
  Embeddings}}.
\newblock In \emph{Advances in Neural Information Processing Systems 29}, pages
  4349--4357. Barcelona, Spain.

\bibitem[{Brants and Franz(2006)}]{brants2006web}
Thorsten Brants and Alex Franz. 2006.
\newblock \href {https://catalog.ldc.upenn.edu/LDC2006T13} {{Web 1T 5-gram
  Version 1}}.
\newblock In \emph{LDC2006T13}, Philadelphia, Pennsylvania. Linguistic Data
  Consortium.

\bibitem[{Cañete et~al.(2020)Cañete, Chaperon, Fuentes, and
  Pérez}]{CaneteCFP2020}
José Cañete, Gabriel Chaperon, Rodrigo Fuentes, and Jorge Pérez. 2020.
\newblock \href {https://pml4dc.github.io/iclr2020/papers/PML4DC2020_10.pdf}
  {{Spanish Pre-Trained BERT Model and Evaluation Data}}.
\newblock In \emph{PML4DC at ICLR 2020}.

\bibitem[{Cocos et~al.(2018)Cocos, Wharton, Pavlick, Apidianaki, and
  Callison-Burch}]{cocos-2018-learning}
Anne Cocos, Skyler Wharton, Ellie Pavlick, Marianna Apidianaki, and Chris
  Callison-Burch. 2018.
\newblock \href {https://doi.org/10.18653/v1/D18-1202} {{Learning Scalar
  Adjective Intensity from Paraphrases}}.
\newblock In \emph{Proceedings of the 2018 Conference on Empirical Methods in
  Natural Language Processing}, pages 1752--1762, Brussels, Belgium.
  Association for Computational Linguistics.

\bibitem[{Devlin et~al.(2019)Devlin, Chang, Lee, and
  Toutanova}]{devlin2019bert}
Jacob Devlin, Ming-Wei Chang, Kenton Lee, and Kristina Toutanova. 2019.
\newblock \href {https://doi.org/10.18653/v1/N19-1423} {{BERT: Pre-training of
  Deep Bidirectional Transformers for Language Understanding}}.
\newblock In \emph{Proceedings of the 2019 Conference of the North {A}merican
  Chapter of the Association for Computational Linguistics: Human Language
  Technologies, Volume 1 (Long and Short Papers)}, pages 4171--4186,
  Minneapolis, Minnesota. Association for Computational Linguistics.

\bibitem[{Fellbaum(1998)}]{Fellbaum1998}
Christiane Fellbaum, editor. 1998.
\newblock \href {http://mitpress.mit.edu/books/wordnet} {\emph{{WordNet: An
  Electronic Lexical Database}}}.
\newblock Language, Speech, and Communication. MIT Press, Cambridge, MA.

\bibitem[{Gar{\'\i}~Soler and Apidianaki(2020)}]{gari2020puntacana}
Aina Gar{\'\i}~Soler and Marianna Apidianaki. 2020.
\newblock \href {https://doi.org/10.18653/v1/2020.emnlp-main.598} {{BERT} knows
  punta cana is not just beautiful, it{'}s gorgeous: Ranking scalar adjectives
  with contextualised representations}.
\newblock In \emph{Proceedings of the 2020 Conference on Empirical Methods in
  Natural Language Processing (EMNLP)}, pages 7371--7385, Online. Association
  for Computational Linguistics.

\bibitem[{Geurts(2010)}]{geurts2010quantity}
Bart Geurts. 2010.
\newblock \emph{Quantity implicatures}.
\newblock Cambridge University Press.

\bibitem[{Grave et~al.(2018)Grave, Bojanowski, Gupta, Joulin, and
  Mikolov}]{grave2018learning}
Edouard Grave, Piotr Bojanowski, Prakhar Gupta, Armand Joulin, and Tomas
  Mikolov. 2018.
\newblock \href {https://www.aclweb.org/anthology/L18-1550} {Learning word
  vectors for 157 languages}.
\newblock In \emph{Proceedings of the Eleventh International Conference on
  Language Resources and Evaluation ({LREC} 2018)}, Miyazaki, Japan. European
  Language Resources Association (ELRA).

\bibitem[{Koutsikakis et~al.(2020)Koutsikakis, Chalkidis, Malakasiotis, and
  Androutsopoulos}]{koutsikakis2020greek}
John Koutsikakis, Ilias Chalkidis, Prodromos Malakasiotis, and Ion
  Androutsopoulos. 2020.
\newblock \href {https://dl.acm.org/doi/10.1145/3411408.3411440} {{GREEK-BERT:
  The Greeks visiting Sesame Street}}.
\newblock In \emph{11th Hellenic Conference on Artificial Intelligence}, pages
  110--117.

\bibitem[{Le et~al.(2020)Le, Vial, Frej, Segonne, Coavoux, Lecouteux, Allauzen,
  Crabb\'{e}, Besacier, and Schwab}]{le2020flaubert}
Hang Le, Lo\"{i}c Vial, Jibril Frej, Vincent Segonne, Maximin Coavoux, Benjamin
  Lecouteux, Alexandre Allauzen, Beno\^{i}t Crabb\'{e}, Laurent Besacier, and
  Didier Schwab. 2020.
\newblock \href {https://www.aclweb.org/anthology/2020.lrec-1.302} {{FlauBERT:
  Unsupervised Language Model Pre-training for French}}.
\newblock In \emph{Proceedings of The 12th Language Resources and Evaluation
  Conference}, pages 2479--2490, Marseille, France. European Language Resources
  Association.

\bibitem[{de~Marneffe et~al.(2010)de~Marneffe, Manning, and
  Potts}]{demarneffe:10}
Marie-Catherine de~Marneffe, Christopher~D. Manning, and Christopher Potts.
  2010.
\newblock \href {https://www.aclweb.org/anthology/P10-1018} {{``Was It Good? It
  Was Provocative.'' Learning the Meaning of Scalar Adjectives"}}.
\newblock In \emph{Proceedings of the 48th Annual Meeting of the Association
  for Computational Linguistics}, pages 167--176, Uppsala, Sweden. Association
  for Computational Linguistics.

\bibitem[{McNally and Boleda(2004)}]{mcnally2004relational}
Louise McNally and Gemma Boleda. 2004.
\newblock \href
  {http://www.cssp.cnrs.fr/eiss5/mcnally-boleda/mcnally-boleda-eiss5.pdf}
  {Relational adjectives as properties of kinds}.
\newblock Colloque de Syntaxe et S{\'e}mantique {\`a} Paris.

\bibitem[{de~Melo and Bansal(2013)}]{demelo:13}
Gerard de~Melo and Mohit Bansal. 2013.
\newblock \href
  {https://tacl2013.cs.columbia.edu/ojs/index.php/tacl/article/view/48} {{Good,
  Great, Excellent: Global Inference of Semantic Intensities}}.
\newblock \emph{Transactions of the Association for Computational Linguistics},
  1:279--290.

\bibitem[{Navigli and Ponzetto(2012)}]{NavigliPonzetto:12aij}
Roberto Navigli and Simone~Paolo Ponzetto. 2012.
\newblock \href
  {http://wwwusers.di.uniroma1.it/~navigli/pubs/AIJ_2012_Navigli_Ponzetto.pdf}
  {{B}abel{N}et: {T}he automatic construction, evaluation and application of a
  wide-coverage multilingual semantic network}.
\newblock \emph{Artificial Intelligence}, 193:217--250.

\bibitem[{Su{\'a}rez et~al.(2019)Su{\'a}rez, Sagot, and
  Romary}]{suarez2019asynchronous}
Pedro Javier~Ortiz Su{\'a}rez, Beno{\^\i}t Sagot, and Laurent Romary. 2019.
\newblock Asynchronous pipeline for processing huge corpora on medium to low
  resource infrastructures.
\newblock In \emph{7th Workshop on the Challenges in the Management of Large
  Corpora (CMLC-7)}. Leibniz-Institut f{\"u}r Deutsche Sprache.

\bibitem[{Taboada et~al.(2011)Taboada, Brooke, Tofiloski, Voll, and
  Stede}]{taboada-etal-2011-lexicon}
Maite Taboada, Julian Brooke, Milan Tofiloski, Kimberly Voll, and Manfred
  Stede. 2011.
\newblock \href {https://doi.org/10.1162/COLI_a_00049} {Lexicon-based methods
  for sentiment analysis}.
\newblock \emph{Computational Linguistics}, 37(2):267--307.

\bibitem[{Wilkinson and Oates(2016)}]{wilkinson2016gold}
Bryan Wilkinson and Tim Oates. 2016.
\newblock \href {https://www.aclweb.org/anthology/L16-1424} {{A Gold Standard
  for Scalar Adjectives}}.
\newblock In \emph{Proceedings of the Tenth International Conference on
  Language Resources and Evaluation ({LREC}'16)}, pages 2669--2675, Portorož,
  Slovenia. European Language Resources Association (ELRA).

\bibitem[{Zipf(1945)}]{Zipf1945}
George~Kingsley Zipf. 1945.
\newblock \href {https://doi.org/10.1006/csla.2001.0174} {The meaning-frequency
  relationship of words}.
\newblock \emph{Journal of General Psychology}, 33(2):251–256.

\end{thebibliography}

\appendix

\begin{table*}[h!]
    \centering
    \scalebox{0.83}{
    \begin{tabular}{cc | ccc | ccc | ccc | ccc}
 
         & \multicolumn{1}{c}{} & \multicolumn{3}{c}{{\sc en}} & \multicolumn{3}{c}{{\sc fr}} & \multicolumn{3}{c}{{\sc es}} & \multicolumn{3}{c}{{\sc el}} \\
  
         & \multicolumn{1}{c}{} & \multicolumn{3}{c}{{\bf Mono WP}} & \multicolumn{3}{c}{{\bf Mono WP}}  & \multicolumn{3}{c}{{\bf Mono WP}}  & \multicolumn{3}{c}{{\bf Mono WP}}\\
         \hline
          & & {\sc p-acc} & $\tau$ & $\rho_{avg}$ & {\sc p-acc} & $\tau$  & $\rho_{avg}$ & {\sc p-acc} & $\tau$ & $\rho_{avg}$ & {\sc p-acc} & $\tau$  & $\rho_{avg}$ \\
          \hline
         \parbox[t]{2mm}{\multirow{2}{*}{\rotatebox[origin=c]{90}{\sc DM 
         }}} & {\sc dv}-1 $(+)$ & \textbf{.664}$_{9}$ & \textbf{.463}$_{9}$ & \textbf{.531}$_{9}$ & \textbf{.617$_{3}$} & \textbf{.384$_{3}$} & \textbf{.406$_{3}$} & \textbf{.652}$_{9}$ & \textbf{.367$_{9}$} & \textbf{.390$_{9}$} & .546$_{8}$ & .201$_{8}$ & .215$_{8}$\\
     
         & {\sc dv-wk} & .557$_{9}$ & .246$_{9}$ & .284$_{6}$ & .517$_{1}$ & .170$_{1}$ & .140$_{1}$ & .645$_{10}$ & .353$_{10}$ & .313$_{10}$ & \textbf{.557}$_{2}$ & \textbf{.226}$_{2}$ & \textbf{.240}$_{2}$\\
         \hline
         \parbox[t]{2mm}{\multirow{2}{*}{\rotatebox[origin=c]{90}{\sc WK
         }}} 
         & {\sc dv}-1 $(+)$ & .852$_{7}$ & .705$_{7}$ & .766$_{1}$ & .612$_{7}$ & .262$_{1}$ & .215$_{6}$ & \textbf{.763}$_{8}$ & \textbf{.525}$_{8}$ & \textbf{.755}$_{6}$ & .632$_{8}$ & .312$_{8}$ & .256$_{8}$\\
 
         & {\sc dv-dm} & \textbf{.918}$_{6}$ & \textbf{.836}$_{6}$ & \textbf{.839}$_{6}$ & .627$_{2}$ & .292$_{2}$ & .392$_{2}$ & .746$_{6}$ & .492$_{6}$ & .658$_{6}$ & \textbf{.779}$_11$ & \textbf{.617}$_{11}$ & \textbf{.663}$_{11}$\\
         \hline
   
   & \multicolumn{1}{c}{} &    
   \multicolumn{3}{|c}{{\bf   Multi WP}} & \multicolumn{3}{|c}{{\bf Multi WP-1}} & \multicolumn{3}{|c}{{\bf Multi WP-1}} & \multicolumn{3}{|c}{{\bf Multi (unc) WP-1}}\\
         \hline
          \parbox[t]{2mm}{\multirow{2}{*}{\rotatebox[origin=c]{90}{\sc DM }}}
          & {\sc dv}-1 $(+)$ & .588$_{4}$ & .301$_{4}$ & .312$_{4}$ & .549$_{7}$ & .239$_{7}$ & .276$_{7}$ & .589$_{3}$ & .229$_{3}$ & .234$_{1}$ & .524$_{9}$ & .153$_{9}$ & .171$_{9}$\\
    
         & {\sc dv-wk} & .516$_{5}$ & .153$_{11}$ & .198$_{5}$ & .490$_{2}$ & .113$_{2}$ & .134$_{7}$ & .603$_{12}$ & .268$_{12}$ & .287$_{12}$ & .521$_{6}$ & .146$_{6}$ & .186$_{6}$\\
         
          \hline

\parbox[t]{2mm}{\multirow{2}{*}{\rotatebox[origin=c]{90}{\sc WK}}} & {\sc dv}-1 $(+)$ & .820$_{7}$ & .639$_{7}$ & .667$_{3}$ & .612$_{3}$ & .262$_{3}$ & .362$_{3}$ & .746$_{4}$ & .492$_{4}$ & .608$_{4}$ & .647$_{9}$ & .358$_{9}$ & .369$_{9}$\\

         & {\sc dv-dm} & .885$_{7}$ & .770$_{7}$ & .834$_{7}$ & \textbf{.687}$_{7}$ & \textbf{.412}$_{7}$ & \textbf{.435}$_{3}$ & .661$_{10}$ & .322$_{10}$ & .447$_{6}$ & .662$_{6}$ & .388$_{6}$ & .444$_{6}$\\
         
    \end{tabular}}
    \caption{Results of {\sc diffvec} ({\sc dv}) methods with contextualised representations derived from monolingual and multilingual models for each language, using an alternative approach to selecting wordpieces (WP, WP-1) than the one used for the results reported in Table \ref{tab:monomulti}. 
    For all languages but Greek, the multilingual model is cased.}
    \label{tab:monomulti_worsewp}
\end{table*}

\section{Comparison of Wordpiece Selection Methods} \label{app:comparison_wp_selection}

Table 3 of the main paper contains  results of the {\sc diffvec} method with the best approach for selecting wordpieces (WPs) for each model. In Table \ref{tab:monomulti_worsewp}, 
we present results obtained 
using the alternative 
approach for each  model and language: 
\begin{itemize}
    \item for all monolingual models and the multilingual model for English, Table \ref{tab:monomulti_worsewp} contains results obtained with the WP approach; 
    \item for the multilingual models in the other languages, we show results with WP-1. 
\end{itemize}

\noindent The best approach was determined by comparing their average scores across the different methods. Some configurations improve, but they yield overall worse results per model, especially in Spanish. 
Differences between WP and WP-1 are generally more pronounced in the multilingual models than in the monolingual models.



\end{document}